\documentclass[runningheads]{llncs}
\usepackage{graphicx}
\usepackage{tikz}
\usepackage{comment}
\usepackage{amsmath,amssymb} 
\usepackage{color}

\usepackage{cite}
\usepackage{epsfig}
\usepackage{amsmath}
\usepackage{amssymb}
\usepackage{booktabs}       
\usepackage{amsfonts}       
\usepackage{nicefrac}       
\usepackage{microtype}      
\usepackage{subfigure}
\usepackage{graphics}
\usepackage{float}
\usepackage{tabularx}
\usepackage{mathrsfs}
\usepackage{enumerate}
\usepackage{multirow}
\usepackage{bbm}
\usepackage{pifont}
\usepackage{lipsum}
\usepackage[colorlinks,linkcolor=blue]{hyperref}

\begin{document}
\pagestyle{headings}
\mainmatter
\def\ECCVSubNumber{2211}  

\title{BorderDet: Border Feature for Dense Object Detection}


\titlerunning{BorderDet: Border Feature for Dense Object Detection}
%
\author{
Han Qiu\inst{1,2*}\and
Yuchen Ma\inst{1*} \and
Zeming Li\inst{1} \and
Songtao Liu\inst{1} \and
Jian Sun\inst{1}
}
\authorrunning{H. qiu, Y. Ma, Z. Li, S. Liu and J. Sun}
%
\institute{
~\inst{1} MEGVII Technology\\
\email{\{qiuhan, mayuchen, lizeming, liusongtao, sunjian\}@megvii.com}\\
~\inst{2} Xi’an Jiaotong University\\
\email{qiuhan@stu.xjtu.edu.cn}\\
}
\maketitle

\begin{abstract}
Dense object detectors rely on the sliding-window paradigm that predicts the object over a regular grid of image. Meanwhile, the feature maps on the point of the grid are adopted to generate the bounding box predictions. The point feature is convenient to use but may lack the explicit border information for accurate localization. In this paper, We propose a simple and efficient operator called Border-Align to extract ``border features'' from the extreme point of the border to enhance the point feature. Based on the BorderAlign, we design a novel detection architecture called BorderDet, which explicitly exploits the border information for stronger classification and more accurate localization. With ResNet-50 backbone, our method improves single-stage detector FCOS by 2.8 AP gains (38.6 v.s. 41.4). With the ResNeXt-101-DCN backbone, our BorderDet obtains 50.3 AP, outperforming the existing state-of-the-art approaches. The code is available at \href{https://github.com/Megvii-BaseDetection/BorderDet}{https://github.com/Megvii-BaseDetection/BorderDet}.

\end{abstract}


\makeatletter
\def\blfootnote{\xdef\@thefnmark{}\@footnotetext}
\makeatother

\blfootnote{
* The first two authors contributed equally to this work.\\
This work was supported in part by the National Key Research and Development Program of China under Grant 2017YFA0700800.
}

\section{Introduction}
\begin{figure}[t]
\centering
\includegraphics[height=0.2204\textwidth, width=0.95\textwidth]{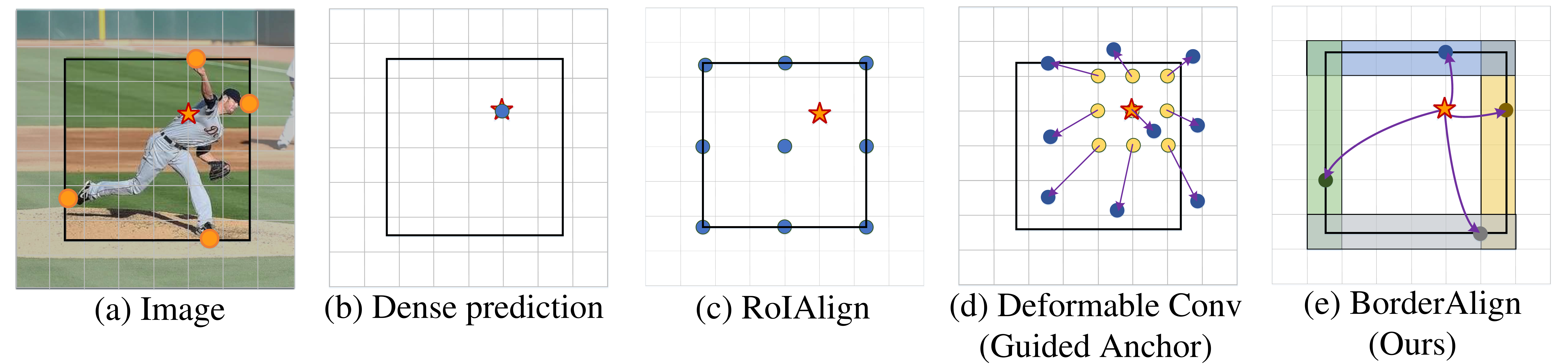}
\caption{Different feature extraction strategy. The red pentagram represents the current point that predicts the bounding box. The black rectangle denotes the bounding box predicted on the red pentagrams. And the blue point indicates where the features are extracted. Different from the deformable Convolution and RoIAlign which densely extract the features from the whole bounding box, Our BorderAlign only extracts the features from five points for the current single point and four extreme points of the borders respectively. The orange points in (a) are the extreme points}
\centering
\label{fig:intro_extractor}
\end{figure}

Sliding-window object detector~\cite{ssd, dssd, yolo, yolo9000, densebox, focal_loss, fcos, single_shot}, which generates bounding-box predictions over a dense and regular grid, plays an essential role in modern object detection. Most sliding-window object detectors like SSD~\cite{ssd}, RetinaNet~\cite{focal_loss} and FCOS~\cite{fcos} adopt a point-based feature representation of the bounding box, where the bounding box is predicted by the feature on each point of the grid, shown as the ``Single Point'' in Fig.~\ref{fig:intro_extractor}. This single point feature is convenient to be used for object localization and object classification because no additional feature extraction is conducted.

However, the point feature may contain insufficient information for representing the full instance with its limited receptive field. Meanwhile, it may also lack the information of the object boundary to precisely regress the bounding box.

Many studies have been focused on the feature representation of the object, such as the GA-RPN~\cite{deformable}, RepPoints~\cite{reppoints} and Cascade RPN~\cite{cascade_rpn}, or pooling based methods like RoI pooling~\cite{fast_rcnn} and RoIAlign~\cite{mask_rcnn}. As shown in Fig.~\ref{fig:intro_extractor}, these methods extract more representative features than the point features. However, there are two limitations of implementing these methods for dense object detection: (1)  The feature extracted 
within the whole boxes may involve \emph{unnecessary} computation and easily be affected by the background. (2) These methods extract the border features \textit{implicitly} and \textit{indirectly}. Since the features are discriminated and extracted adaptively within the whole boxes, no specific extraction on the border features is conducted in these methods.

In this work, we propose a powerful feature extraction operator called BorderAlign, which directly utilizes the border features pooled from each boundary to enhance the original point feature. It differs from the other feature extraction operators as shown in Fig.~\ref{fig:intro_extractor}, which densely extracts the feature from the whole box. Our proposed BorderAlign focuses on the object border and is designed to adaptively discriminate the representative part of the object border, \emph{e.g}. the extreme point~\cite{extremenet}, which is shown in Fig.~\ref{fig:intro_extractor}(e).

We design BorderDet which utilizes Border Alignment Modules~(BAM) to refine the classification score and bounding box regression. Our BorderDet uses less computation than similar feature enhancement methods and achieves better accuracy. Moreover, our method can be easily integrated into any dense object detectors with/without anchors.

To summarize, our contribution is three-fold as follows:
\begin{enumerate}[1]
\item We analyze the feature representation for the dense object detector and demonstrate the significance of supplementing the single-point feature representation with the border feature.
\item We propose a novel feature extraction operator called BorderAlign to enhance features by the border features. Based on BorderAlign, we present an efficient and accurate object detector architecture named BorderDet.
\item We achieve state-of-the-art results on COCO dataset without bells and whistles. Our method leads to significant improvements on both single-stage method FCOS and two-stage method FPN, by 2.8 $AP$ and 3.6 $AP$ respectively. Our ResNext-101-DCN based BorderDet yields 50.3 $AP$, outperforming the existing state-of-the-art approaches.
\end{enumerate}

\section{Related Works}
\subsubsection{\textbf{Sliding-window Paradigm.}}
Sliding-window Paradigm is widely used in object detection. For the one-stage object detectors, Densebox~\cite{densebox}, YOLO~\cite{yolo,yolo9000}, SSD~\cite{ssd}, RetinaNet~\cite{focal_loss}, and FCOS~\cite{fcos} have demonstrated the effectiveness to densely predict the classification and localization scores.
For the two-stage object Detectors, R-CNN series~(\cite{girshick2014rcnn,sppnet, rfcn,fast_rcnn,faster_rcnn,fpn,mask_rcnn}) adopt the region proposal network~(RPN) that based on the sliding-window mechanism to generate the initial proposals, and then a refinement stage that consists of a RoIAlign~\cite{mask_rcnn} and R-CNN is performed to warp the feature maps of the region-of-interests (RoI) and generate the accurate predictions.

\subsubsection{\textbf{Feature Representation of Object.}}
Typical sliding-window object detectors adopt a point-based feature representation. However, it is difficult for the point feature to maintain the powerful feature representation for both classification and location.  Recently, some works~\cite{guided_anchor, cascade_rpn, reppoints} attempt to improve the feature representation of object detection. Guided Anchor~\cite{guided_anchor} is proposed to enhance the single point feature representation by utilizing the deformable convolution. Cascade-RPN~\cite{cascade_rpn} presents adaptive convolution to align the features maps to their corresponding object bounding box predictions. Reppoints~\cite{reppoints} formulate the object bounding box as a set of representative points and extract the representative point feature by deformable convolution. However, the feature maps proposed in these methods are extracted from the whole object, thus the feature extraction is redundant and easily affected by the background feature maps. In contrast to the above methods, our BorderDet directly enhances the single point feature by the border feature, which enables the feature map to have a high response to the extreme points of the object borders and does not involve the background noise.

\subsubsection{\textbf{Border Localization.}}
There are several methods that search on each row and column of the region or bucket to accurately locate the boundary of the object. LocNet~\cite{locnet} and SABL~\cite{sabl} adopt an additional object localization stage which aggregates the RoI feature maps along with X-axis and Y-axis to locate the object borders and generate the probability for each object border prediction. However, such border localization pipelines rely heavily on the high-resolution RoI feature maps, thus the implementation of these methods in the dense object detector may be restricted. In this work, we aim to efficiently exploit the border feature for accurate object localization.


\section{Our Approach}
In this section, we first investigate the feature representation of the bounding box in sliding-window object detectors. Then, we propose a new feature extractor called BorderAlign, which extracts the border features to enhance the original point-based feature representation. Based on the BorderAlign, we present the design of BorderDet and discuss its mechanism for extracting boundary features efficiently.

\begin{figure}[t]
\centering
\includegraphics[height=0.208\linewidth, width=1.0\linewidth]{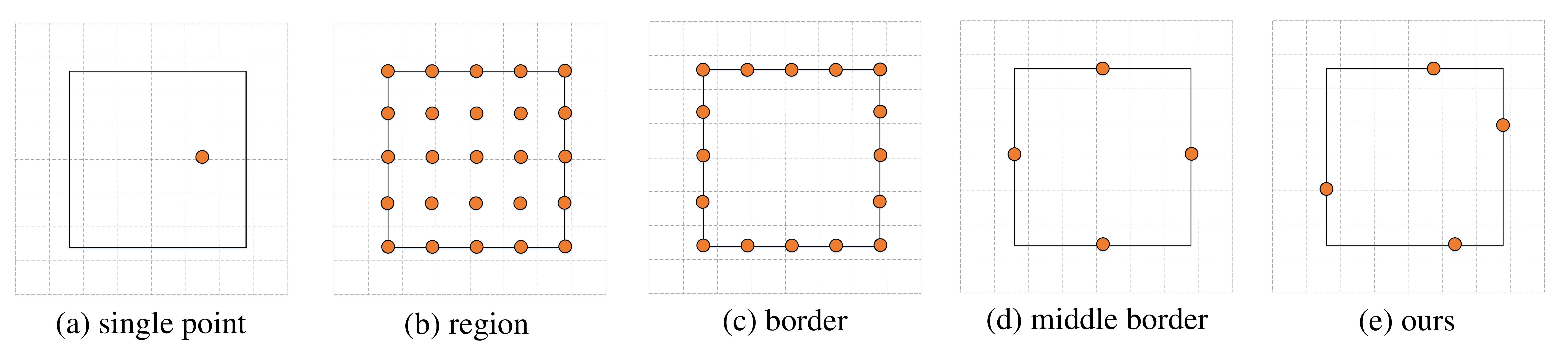}
\caption{Different feature representations of the bounding box. (a) denotes \emph{point feature} on each point of the grid. (b) indicates the \emph{region features} extracted from the whole bounding box by RoIAlign~\cite{mask_rcnn}. (c) denotes the \emph{border features} extracted from the border of the bounding box. (d) indicates the \emph{border-middle features} which are extracted from the center point of each border. (e) denotes our BorderAlign feature extractor}
\centering
\label{fig:motivation}
\end{figure}

\subsection{Motivation}
Sliding-window object detectors usually generate bounding box predictions over a dense, regular grid of feature maps. As shown in Fig.~\ref{fig:motivation}, the feature on each point of the grid is generally used to predict the category and location of the objects. This point-based feature representation is hard to contain the effective border feature and it may limit the localization ability of the object detectors. As for the two-stage object detectors, the object is described by the region features which are extracted from the whole bounding box, which is shown in Fig.~\ref{fig:motivation} (b). This region-based feature representation is able to provide more abundant features than the point-based feature representation for object classification and localization.

In Table.~\ref{tab:motivation}, we provide a deeper analysis of the feature representation of the bounding box. Firstly, we adopt a simple dense object detector(FCOS) as our baseline to generate the coarse bounding box predictions. Next, we will re-extract the features as shown in Fig.~\ref{fig:motivation} from the second-to-last feature map of FCOS. Then we gradually supplement the single point feature with different features to refine the coarse predictions. We make the following observations on these experiments. (1) Region features are more representative than the point feature. Enhancing the single point feature with the region features leads to an improvement of 1.3 AP. (2) The border features play a major role in the region features when the region features are used to enhance the single point feature. Performance only reduces 0.3 AP if we ignore the inner part of the bounding box and only introduce the border features. (3) Extracting the border features effectively leads to further improvement than densely extracting the border features. The experiment in the fourth column of Table.~\ref{tab:motivation} shows that the middle border features is 0.3 AP higher than border features and reaches the same performance to the region features with fewer sample points.

\setlength{\tabcolsep}{2.8pt}
\begin{table}[t]
\caption{Comparison of different feature representation of the bounding box. The first row is the baseline. The $F_{point}$ indicates the feature used in the first prediction. $F_{point}^{'}$, $F_{region}$, $F_{border}$ and $F_{middle}$ indicate the features used in the second prediction. The specific illustration of these features are shown in Fig.~\ref{fig:motivation}. The final column ``N'' denotes how many points are sampled to extract feature in the second prediction where ``N'' equals 5 in these experiments}
\centering
\begin{tabular}{c|cccc|cccccc|c}
\toprule
$F_{point}$ & $F_{point}^{'}$ & $F_{region}$ & $F_{border}$ & $F_{middle}$ & $AP$ & $AP_{50}$ & $AP_{75}$ & $AP_{S}$ & $AP_{M}$ & $AP_{L}$ & $N$ \\
\midrule
\checkmark &            &            & & & 38.6 & 57.2 & 41.7 & 23.5 & 42.8 & 48.9 & 0 \\
\midrule
\checkmark & \checkmark &            & & & 38.9 & 57.7 & 42.1 & 23.7 & 43.1 & 49.3 & $1$  \\
\checkmark & \checkmark & \checkmark & & & \textbf{39.9} & 58.9 & 43.4 & 24.6 & 44.1 & 50.8 & $n^{2}+1$ \\
\checkmark & \checkmark & & \checkmark & & 39.6 & 58.5 & 43.2 & 24.2 & 43.8 & 50.4 & $4n+1$\\
\checkmark & \checkmark & & & \checkmark & \textbf{39.9} & 58.7 & 43.4 & 24.8 & 44.0 & 50.4 & $4+1$ \\
\bottomrule
\end{tabular}
\label{tab:motivation}
\end{table}

In consequence, for the feature representation in the dense object detector, the point-based feature representation is lack of the explicit feature of the whole object and the feature enhancement is requisite. However, extracting the feature from the whole boxes is unnecessary and redundant. Meanwhile, a more efficient extraction strategy of the border features will lead to better performance. Based on these conceptions, we explore how to boost the dense object detector performance by using border feature enhancement in the next section.

\begin{figure}[t]
\centering
\includegraphics[height=0.432\linewidth, width=1.0\linewidth]{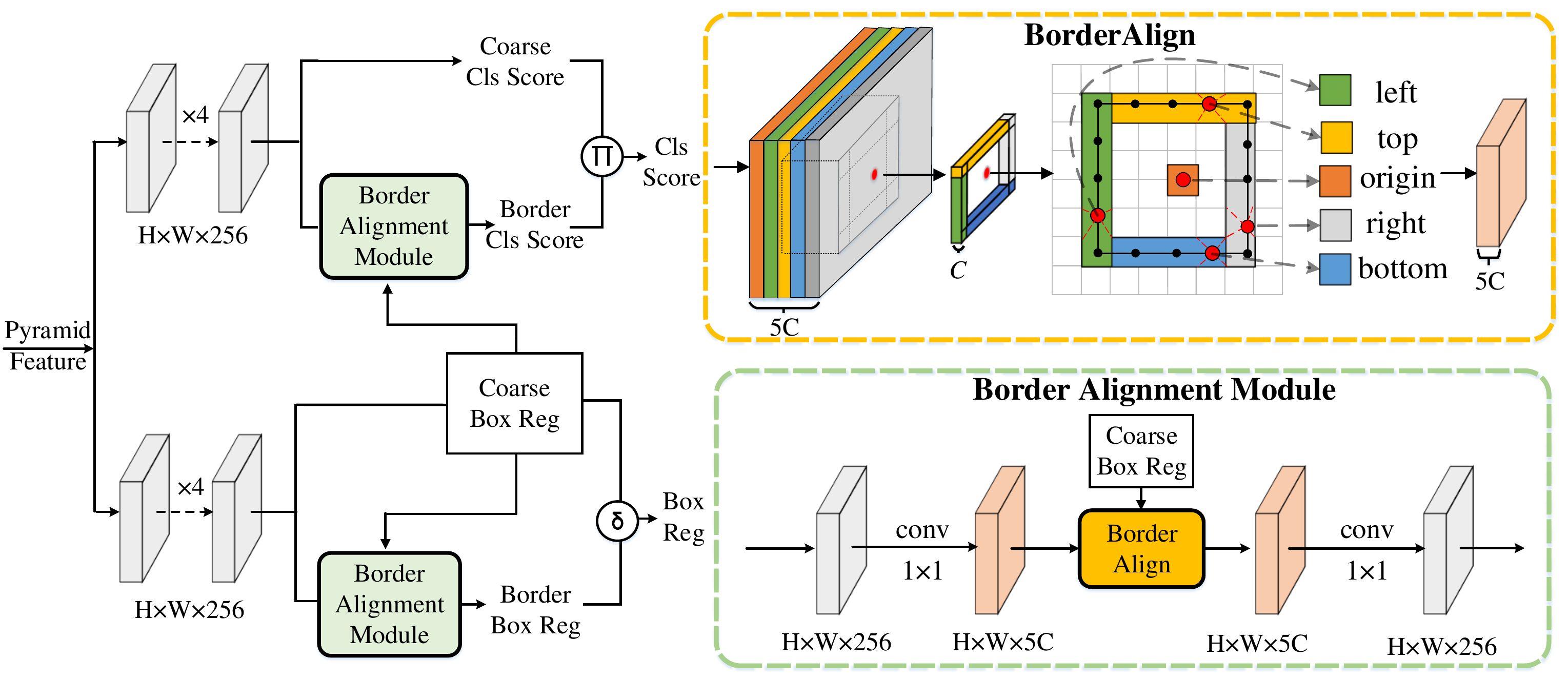}
\caption{The architecture of BorderDet. Firstly, we adopt a regular single-stage object detector to generate the coarse predictions of the classification score and bounding box location. Then the Border Alignment Module is applied to refine the coarse predictions with the border features. The $\pi$ indicates multiplication and the $\delta$ denotes the combination of the two bounding box locations}
\label{fig:architecture}
\end{figure}

\subsection{Border Align}
\label{sec:BorderAlign}
Owing to our observation above, the border features are important in achieving better detection performance. However, it is inefficient to extract features intensively on the borders since there is usually very little foreground and lots of background on the borders of the object(\emph{e.g.} the person in Fig.~\ref{fig:intro_extractor}). We thus propose a novel feature extractor, called BorderAlign to effectively exploit the border feature.

The architecture of the BorderAlign is illustrated in Fig.~\ref{fig:architecture}. Inspired by the R-FCN~\cite{rfcn}, our BorderAlign take the \textit{border-sensitive} feature maps $I$ with $(4+1)C$ channels as the input. The $4C$ channels of the feature maps correspond to the four borders (left, top, right, bottom), while the other $C$ channels corresponds to the original single point features as shown in Fig.~\ref{fig:motivation}. Then, each border is evenly subdivided into N points and the feature values of these N points are aggregated by the max-pooling. N denotes the pooling size and is set to 10 in this paper as default. The proposed BorderAlign could adaptively exploit the representative border features from the extreme points of the borders.

It is worth noting that our BorderAlign adopts a channel-wise max-pooling scheme that the four borders are max-pooled independently within each $C$ channels of the input feature maps. Assuming the input feature maps are in the order of (single point, left border, top border, right border and bottom border), the output feature maps $\mathcal{F}$ can be formulated as the following equation:

\begin{equation}
\small
    F_{c}(i,j) = \left\{
    \begin{array}{ll}
        I_{c}(i, j)                                                   \quad \quad & \quad 0\le c<C    \\
        \max\limits_{0\le k\le N-1}(I_{c}(x_{0}, y_{0}+\frac{kh}{N})) \quad \quad & \quad C\le c<2C    \\
        \max\limits_{0\le k\le N-1}(I_{c}(x_{0}+\frac{kw}{N}, y_{0})) \quad \quad & \quad 2C\le c< 3C  \\
        \max\limits_{0\le k\le N-1}(I_{c}(x_{1}, y_{0}+\frac{kh}{N})) \quad \quad & \quad 3C\le c< 4C  \\
        \max\limits_{0\le k\le N-1}(I_{c}(x_{0}+\frac{kw}{N}, y_{1})) \quad \quad & \quad 4C\le c< 5C  \\
    \end{array}
    \right.
\end{equation}

Here $\mathcal{F}_{c}(i,j)$ is the feature value on the $(i,j)$-th point for the $c$-th channel of the output feature maps $\mathcal{F}$, $(x_{0},y_{0},x_{1},y_{1})$ is the bounding box prediction on the point $(i,j)$, $w$ and $h$ are the width and height of $(x_{0},y_{0},x_{1},y_{1})$. To avoid the quantization error, the exact value $I_{c}$ is computed by bilinear interpolation~\cite{Spatial_transformer_networks} with the nearby feature value on the feature maps.

\begin{figure}[t]
\centering
\includegraphics[height=0.4\linewidth, width=0.7\linewidth]{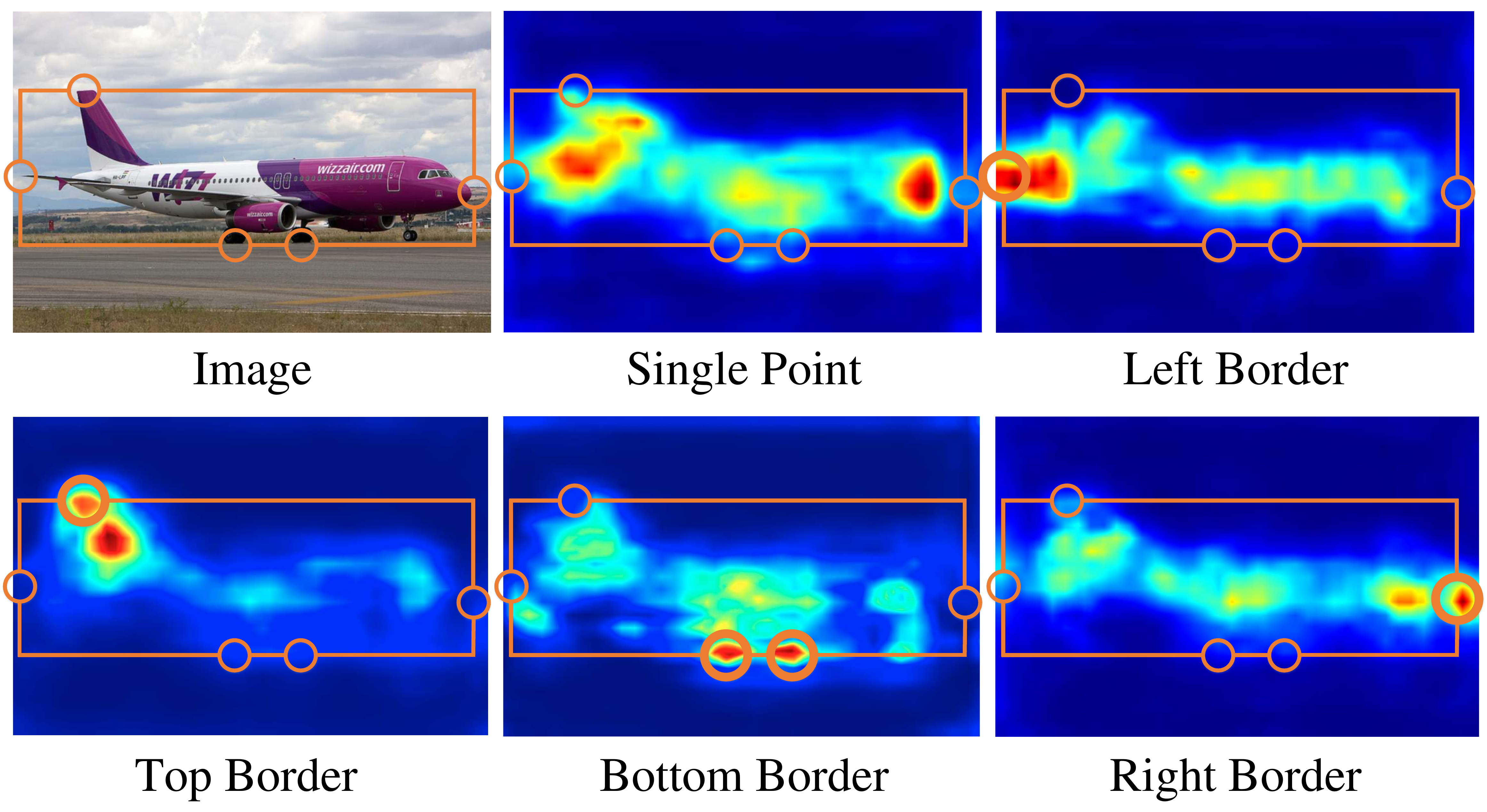}
\caption{Visualization of the \textit{border-sensitive} feature maps. The orange circle on the border indicate the extreme points. The feature maps of 'Single Point', 'Left Border', 'Top Border', 'Right Border' and 'Bottom Border' are the maximum feature value on each $C$ channels of the \textit{border-sensitive} feature maps}
\label{fig:border image}
\end{figure}

In Figure.~\ref{fig:border image}, we visualize the maximum value on each $C$ channels of the \textit{border-sensitive} feature maps. It reveals that the bank of $(4+1)C$ feature maps are guided to activated in their corresponding location of the object. For example, the first $C$ channels of the show strong response over the whole object. Meanwhile, the second $C$ exhibits a high response near the left border of the object. These \textit{border-sensitive} feature maps facilitate our BorderAlign to extract the border feature in a principle way.

\subsection{Network Architecture}

\label{sec:BorderDet}
\subsubsection{\textbf{BorderDet.}}  We now present the network architecture of our BorderDet. In our experiments, we adopt a simple anchor-free object detector FCOS as our baseline. Since the border extraction procedure in BorderAlign requires border location as input, our BorderDet adopts two prediction stages as shown in Fig.~\ref{fig:architecture}. Taken the pyramid feature maps as input, the BorderDet first predicts the coarse classification scores and coarse bounding box locations. Then the coarse bounding box locations and the feature maps are fed into the Border Alignment Module~(BAM) to generate the feature maps which contain explicit border information. Finally, we apply a $1\times 1$ convolutional layers to predict the border classification score and border locations. The above two predictions will be unified to form the final predictions. It is to note that the border classification score is category-aware to avoid ambiguous predictions when there is an overlapping among different category boundaries.



It is worth noting that although our BorderDet adopts two extra predictions for both object classification and object localization, the additional computation is negligible due to the effective structure and layer sharing. In addition, the proposed method can be integrated into other object detectors in a plug-and-play manner, including RetinaNet~\cite{focal_loss}, FCOS~\cite{fcos} and so forth.

\label{BAM}
\subsubsection{\textbf{Border Alignment Module.}} The structure of Border Alignment Module~(BAM) is illustrated in the green box of Fig.~\ref{fig:architecture}. BAM takes the feature maps with $C$ channels as input, followed by a $1\times 1$ convolutional layer with instance normalization to output the \textit{border-sensitive} feature maps. The border-sensitive feature maps composed of five feature maps with $C$ channels for each border and the single point. Thus the channels of the output feature maps have $(4+1)C$ channels. In our experiments, $C$ is set to 256 for the classification branch and to 128 for the regression branch. Finally, we adopt BorderAlign module to extract the border feature from the \textit{border-sensitive} feature maps, and apply a $1\times 1$ convolutional layer to reduce the $(4+1)C$ channels back to $C$.

\label{BorderRPN}
\subsubsection{\textbf{BorderRPN.}} 
\begin{figure}[t]
\centering
\includegraphics[height=0.24\linewidth, width=0.83\linewidth]{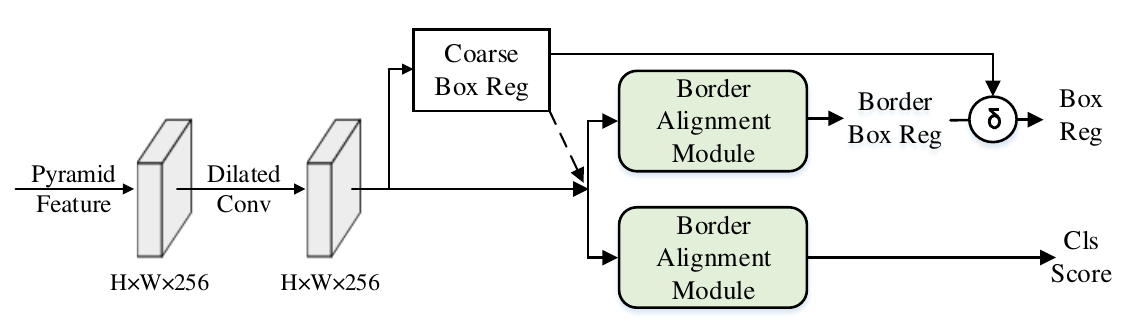}
\caption{The architecture of BorderRPN. We use the BAM to enhance the origin feature of the RPN, and combine the coarse bounding box locations and the border locations by $\delta$. Meanwhile, we use the border classification score as the classification score of BorderRPN}
\label{fig:borderrpn}
\end{figure}

Our method can also be served as a better proposal generator for the typical two-stage detectors. We add the border alignment module to RPN and denote the new structure as BorderRPN. The architecture of BorderRPN is shown in Fig.~\ref{fig:borderrpn}. We remain the regression branch in RPN to predict the coarse bounding box locations. The first $3\times3$ convolution in RPN is replaced with a $3\times3$ dilated convolution to increase the effective receptive field.

\subsection{Model Training and Inference}
\textbf{Target Assignment}. We adopt FCOS~\cite{fcos} as our baseline to predict the coarse classification scores and coarse bounding box prediction $(x_{0},y_{0},x_{1},y_{1})$. Then, in the second stage, a coarse bounding box prediction will be assigned to the ground-truth box $(x_{0}^{t},y_{0}^{t},x_{1}^{t},y_{1}^{t})$ by using an intersection-over-union~(IoU) threshold of 0.6. And its regression targets $(\delta{x}_{0},\delta{y}_{0},\delta{x}_{1},\delta{y}_{1})$ are computed as following:

\begin{equation}
\normalsize
\begin{array}{lr}
\delta{x}_{0}=\frac{x_{0}^{t}-x_{0}}{w*\sigma} \quad \delta{y}_{0}=\frac{y_{0}^{t}-y_{0}}{h*\sigma} \quad
\delta{x}_{1}=\frac{x_{1}^{t}-x_{1}}{w*\sigma} \quad \delta{y}_{1}=\frac{y_{1}^{t}-y_{1}}{h*\sigma} ,
\end{array}
\end{equation}
where $w$, $h$ are the width and height of the coarse bounding box prediction, and $\sigma$ is the variance to improve the effectiveness of multi-task learning~($\sigma$ equals to 0.5 by default).

\textbf{Loss Function.} The proposed BorderDet is easy to optimize in an end-to-end way using a multi-task loss. Combining the output of the BorderDet, we define our training loss function as follows:

\begin{equation}
\begin{aligned}
\begin{split}
\mathcal{L}= & \mathcal{L}^C_{cls} + \mathcal{L}^C_{reg} + \frac{1}{\mathcal{N}_{pos}}\sum_{x,y}\mathcal{L}_{cls}^B(\mathcal{P}^B,\mathcal{C}^{*}) + \mathcal{L}_{reg_\{\mathcal{C}^{*}>0\}}^B(\Delta,\Delta^{*}), \\
\end{split}
\end{aligned}
\end{equation}


\noindent where $\mathcal{L}_{cls}^{C}$ and $\mathcal{L}_{reg}^{C}$ are the coarse classification loss and coarse regression loss. In the implementation, focal loss~\cite{focal_loss} and IoU loss are used as the classification loss and regression loss respectively, which are the same as FCOS~\cite{fcos}. $\mathcal{L}_{cls}^{B}$ is the focal loss computed between the border classification and its assigned ground truth $\mathcal{C}^{*}$, the loss is averaged by the number of the positive
samples ${\mathcal{N}_{pos}}$. We use $\mathcal{L}_{1}$ loss as our corner regression loss. $\mathcal{P}^{B}$ represents the predicted border classification scores and $\Delta$ is the predicted border offset.

\textbf{Inference.} BorderDet predicts classification scores and box locations for each pixel on the feature maps, while the final classification score is obtained by multiplying the coarse score and border score. The bounding boxe location is computed in a simple transformation as illustrated above. Finally, the predictions from all levels are merged and non-maximum suppression(NMS) with a threshold of 0.6.

\section{Experiments}

\subsection{Implementation Details}
\label{sec:exp_setting}
Following the common practice, our ablation experiments are trained on COCO trainval35k set (115K images) and evaluated on COCO minival set (5K images). To compare with the state-of-art approaches, we report COCO AP on the test-dev set (20K images).
We use ResNet-50 with FPN as our backbone network for all the experiments, if not otherwise specified. We use synchronized stochastic gradient descent(SGD) over 8 GPUs with a total of 16 images per minibatch (2 images per GPU) for 90k iterations. With an initial learning rate of 0.01, we decrease it by a factor of 10 after 60k iterations and 80k iterations respectively. We use horizontal image flipping as the only form of data augmentation. Weight decay of 0.0001 and momentum of 0.9 are used. We initialize our backbone network with the weights pre-trained on ImageNet. Unless specified, the input images are resized to ensure their shorter edge being 800 and the longer edge less than 1333.

\subsection{Ablation Study}

We gradually add the Border Alignment Module~(BAM) to the baseline to investigate the effectiveness of our proposed BorderDet. We first apply the BAM on the classification branch. As shown in the second row of Table.~\ref{tab:ablation study}, the BAM leads to a gain of 1.1 AP. It is worth noting that the improvement mainly occurs in the AP with a low threshold and the improvement decays along with the increase of IoU threshold. The improvement at low IoU threshold is because the BAM can rescore the bounding boxes according to their border features, and maintain the predictions with both high classification score and localization accuracy. And the performance at a high IoU threshold is restricted by a lack of high-quality detected bounding boxes.

As opposed to BAM on the classification branch, the improvement made by the BAM on the regression branch is mainly concentrated on the AP with the high IoU threshold. The third row in Table.~\ref{tab:ablation study} shows that conducting the BAM on the regression branch boosts the performance from 38.6 to 39.7. The BAM on the regression branch can significantly raise the localization accuracy of the detected bounding boxes, and lead to a gain of 2.6 $AP_{90}$.

Finally, as shown in the last row in Table.~\ref{tab:ablation study}, the implementation of the BAM on both branches can further improve the performance from 38.6 to 41.4. And the improvements are achieved over all IoU thresholds~(from $AP_{50}$ to $AP_{90}$), with $AP_{50}$ increasing by 2.2 and $AP_{90}$ by 3.5. It is worth mentioning that the $AP_{90}$ has been improved by 20\% compared with the baseline. This dramatic performance improvement demonstrates the effectiveness of our proposed BorderDet, especially for the detection with high IoU thresholds.

\begin{table}[t]
\setlength{\tabcolsep}{4pt}
\caption{Ablation studies of BorderDet. The 'cls' and 'reg' denote the implementation of the Border Alignment Module (BAM) on the classification branch and the regression branch respectively}
\centering
\begin{tabular}{c|c|cccccc}
\toprule
Cls-BAM  & Reg-BAM  & $AP$ & $AP_{50}$ & $AP_{60}$ & $AP_{70}$ & $AP_{80}$ & $AP_{90}$\\
\midrule
           &            & 38.6 & 57.2 & 53.3 & 46.7 & 35.3 & 16.0 \\
\midrule
\checkmark &            & 39.7 & 58.4 & 54.8 & 48.5 & 36.2 & 15.9 \\
           & \checkmark & 39.7 & 57.3 & 53.3 & 47.3 & 36.9 & 18.6 \\
\checkmark & \checkmark & \textbf{41.4} & \textbf{59.4} & \textbf{55.4} & \textbf{49.4} & \textbf{38.6} & \textbf{19.5} \\
\bottomrule
\end{tabular}
\label{tab:ablation study}
\end{table}

\begin{table}[t]
\setlength{\tabcolsep}{6pt}
\caption{Ablation studies on the pooling size of the BorderAlign. Considering the speed/accuracy trade-off, pooling size equals to 10 in all our experiments}
\centering
\begin{tabular}{c|cccccccc}
\toprule
Size  & 0    &   2   & 4     & 6     & 8     & 10    & 16    & 32    \\
\midrule
$AP$  & 39.0 & 40.5  & 41.2  & 40.9  & 41.0  & \textbf{41.4} & 41.3  & 41.4  \\
$fps$ & 18.3 & 17.0  & 17.0  & 16.9  & 16.9  & 16.7  & 16.6  & 15.9  \\
\bottomrule
\end{tabular}
\label{tab:pooling size}
\end{table}

\subsection{Border Align}
\noindent\textbf{Pooling Size.} As described in Sec.~\ref{sec:BorderAlign}, the BorderAlign first subdivides each border into several points and then pools over each border to extract the border features. One new hyper-parameter, the pooling size, is introduced during the procedure of BorderAlign. We compare the detection performance of different pooling size in BorderAlign. Results are shown in Table.~\ref{tab:pooling size}. When the pooling size equals $0$, the experiment is equivalent to iteratively predict the bounding box. The experiments show that the results are robust to the value of pooling size in a large range. As a large pooling size expenses extra computation, while a small pooling size leads to unstable results, the pooling size is set to 10 as our default setting.

\noindent\textbf{Border-Sensitive Feature Maps.} To analyse the impact of the \textit{border-sensitive} feature maps as described in Sec.~\ref{sec:BorderAlign}, we also apply the BorderAlign on border-agnostic feature maps with $C$ channels. All the features in BorderAlign will be extracted from the same $C$ feature maps. As shown in Table.~\ref{tab:Border-Sensitive Feature Maps}, \textit{border-sensitive} feature maps improve the $AP$ from 40.8 to 41.4. This is because the \textit{border-sensitive} feature maps could be highly activated on the extreme points of different borders on different channels, thus facilitate the border feature extraction.

\setlength{\tabcolsep}{6pt}
\begin{table}[t]
\caption{Ablation studies on the border-sensitive feature maps. `border sensitive' indicates that the extractions of border features and original single point feature are conducted on the different feature maps, while `border agnostic' means the feature extractions are conducted on the single feature map}
\centering
\begin{tabular}{c|cccccc}
\toprule
& $AP$ & $AP_{50}$ & $AP_{75}$ & $AP_{S}$ & $AP_{M}$ & $AP_{L}$\\
\midrule
border agnostic  & 40.8 & 59.1 & 44.0 & 23.7 & 44.7 & 52.7 \\
border sensitive & 41.4 & 59.4 & 44.5 & 23.6 & 45.1 & 54.6 \\
\bottomrule
\end{tabular}
\label{tab:Border-Sensitive Feature Maps}
\end{table}

\begin{table}[t]
\caption{Ablation studies on border feature aggregation strategy in BorderAlign. For the ``Border-Wise'' strategy, firstly, the feature maps are aggregate along channel dimension by different pooling methods to generate the feature maps with channel equals $1$. Then, a max-pooling is conducted on each border of the object to explore the extreme point, and the feature maps on the extreme points are extracted to form the border features. For the ``Channel-Wise'' strategy, the border feature of each channel is aggregated along the border by average-pooling or max-pooling independently}
\centering
\begin{tabular}{c|c|cccccc}
\toprule
\multicolumn{2}{c|}{Aggregation Strategy} & $AP$ & $AP_{50}$ & $AP_{75}$ & $AP_{S}$ & $AP_{M}$ & $AP_{L}$\\
\midrule
\multirow{2}*{Border-Wise} & average-pooling & 39.9 & 58.8 & 43.0 & 23.0 & 43.8 & 51.9\\
                           & max-pooling & 39.5 & 58.2 & 42.7 & 22.6 & 43.3 & 51.3\\
\midrule
\multirow{2}*{Channel-Wise} & average-pooling & 40.6 & 58.9 & 43.8 & 23.8 & 44.4 & 52.7\\
                    & max-pooling & \textbf{41.4} & 59.4 & 44.5 & 23.6 & 45.1 & 54.6 \\

\bottomrule
\end{tabular}
\label{tab:feature aggregate strategy}
\end{table}

\begin{table}[t]
\small
\caption{Comparison of different feature extraction strategies. All the fps of the extraction strategies are tested on a single NVIDIA 2080Ti GPU}
\centering
\begin{tabular}{l|cccccc|c}
\toprule
Method & $AP$ & $AP_{50}$ & $AP_{75}$ & $AP_{S}$ & $AP_{M}$ & $AP_{L}$ & fps   \\
\midrule
FCOS~\cite{fcos}                    & 38.6 & 57.2 & 41.7 & 23.5 & 42.8 & 48.9 & 18.4 \\
\midrule
w/Iter-Box~\cite{iterative_refine}  & 39.0 & 58.0 & 42.0 & 21.8 & 42.9 & 50.7 & 18.3 \\
w/Adaptive Conv~\cite{cascade_rpn}  & 39.6 & 58.5 & 42.8 & 22.0 & 43.5 & 51.3 & 16.8 \\
w/Deformable Conv~\cite{deformable} & 39.5 & 58.5 & 42.9 & 22.0 & 43.5 & 52.0 & 16.8 \\
w/RoIAlign~\cite{mask_rcnn}         & 40.4 & 58.6 & 43.6 & 22.6 & 44.1 & 53.1 & 12.6       \\
w/\textbf{BorderAlign}              & \textbf{41.4} & \textbf{59.4} & \textbf{44.5} & \textbf{23.6} & \textbf{45.1} & \textbf{54.6} & 16.7  \\
\bottomrule
\end{tabular}
\label{tab:comparison of feature alignment}
\end{table}

\noindent\textbf{Border Feature Aggregation Strategy.} In BorderAlign, we adopt a channel-wise max-pooling strategy that the border feature of each channel is aggregated along the border independently. We investigate the influence of the aggregation strategy from both the channel-wise and border-wise. As illustrated in Table.~\ref{tab:feature aggregate strategy}, the channel-wise max-pooling strategy achieves the best performance of 41.4 $AP$. Compare to the other methods, the proposed channel-wise max-pooling strategy could extract the representative border feature without involving the background noise.

\noindent\textbf{Comparison with Other Feature Extraction Operators.} Cascade-RPN~\cite{cascade_rpn} and GA-RPN~\cite{guided_anchor} proposed to ease the misalignment between the prediction bounding boxes and their corresponding feature. Both the two methods adopt some irregular convolutions, like deformable convolution~\cite{deformable} and adaptive convolution~\cite{cascade_rpn} to extract the feature of the bounding boxes. These irregular convolutions can also extract the border feature implicitly. To further prove the effectiveness of our proposed BorderAlign, we directly replace the BorderAlign and the second convolution in Border Alignment Module (BAM) (Fig.~\ref{fig:architecture}) with the adaptive convolution and deformable convolution respectively. For the fair comparison, we remain the first $1\times1$ convolution with Instance Nornalization~\cite{instancenorm} in the BorderDet. Meanwhile, we also compare the BorderAlign with the RoIAlign by replacing the BorderAlign with RoIAlign. Table.~\ref{tab:comparison of feature alignment} reveals that the BorderAlign outperforms other feature extraction operators by 1.0 AP at least.

Our proposed BorderAlign can concentrate on the representative part of the border, like the extreme points, and extract the border features explicitly and efficiently. On the contrary, the other operators which extract the feature from the whole boxes will introduce the redundant features and limit the detection performance.

\begin{figure}[t]
\centering
\begin{minipage}[t]{1.0\textwidth}
\includegraphics[height=0.361\textwidth, width=0.95\textwidth]{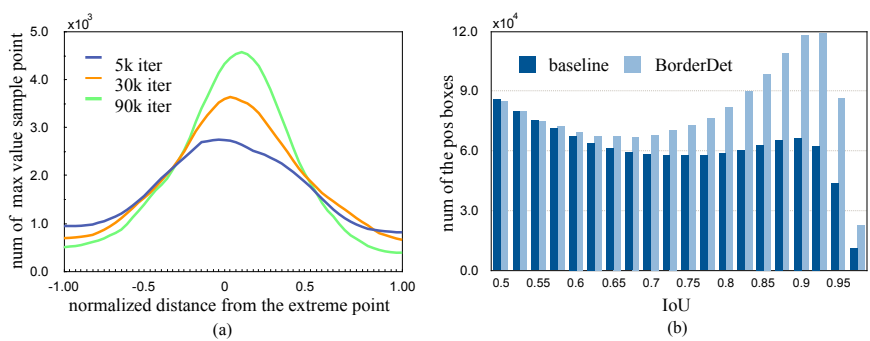}
\end{minipage}
\caption{(a) The statistical analysis of the border extraction. The horizontal axis indicates the normalized distance from the extreme point to the point with max feature value in BorderAlign. (b) The IoU histogram of output bounding boxes. Both the quality and quantity of the output boxes have been greatly improved by the BorderDet}
\label{fig:analysis}
\end{figure}

\subsection{Analysis of BorderDet}

\textbf{Border Feature Representation.}
BorderAlign is accomplished by a channel-wise max-pooling along the border that guarantees the feature extraction process is conducted around the representative extreme points on the borders. We demonstrate this perspective by a quantitative method. Concretely, we first use the annotations of the instance segmentation to yield the locations of the extreme points (top-most, left-most, bottom-most, right-most). Then, we calculate the counts of the normalized distances from the BorderAlign sample points to the extreme points in each response maps during the training(5k iteration, 30k iterations and 90k iterations), which is shown in Fig.\ref{fig:analysis}. The mean value of the normalized distance is almost equal to zero. Meanwhile, the variance of the distance decreased gradually during the training. It means that our BorderDet can adaptively learn to extract the feature near the extreme point. These results further demonstrate the effectiveness of the proposed BorderDet for border feature extraction.

\textbf{Regression Performance.}
To further investigate the benefit of the border feature in object localization, we count the number of bounding box predictions with different IoU thresholds separately. Fig.~\ref{fig:analysis} shows the comparison of the distributions of the bounding box predictions in the FCOS and BorderDet. We can see the localization accuracy of the bounding boxes is improved significantly. The number of valid prediction boxes (IoU greater than 0.5) increased by about 30$\%$. In particular, the number of boxes with IoU greater than 0.9 has nearly doubled. This observation can also explain the significant improvement in $AP_{90}$ as shown in Table.~\ref{tab:ablation study}.


\subsection{Generalization of BorderDet}
Our BorderDet can be easily integrated with the many popular object detectors, e.g. RetinaNet and FPN. To prove the generalization of the BorderDet, we first add the proposed border alignment module to the RetinaNet. For a fair comparison, without modifying any setting of RetinaNet, we directly select the one with the highest score from the nine prediction boxes of each pixel to refine. As shown in Table~\ref{tab:generalization of BorderDet}, BorderDet can consistently improve the RetinaNet by 2.3 $AP$. For the two-stage method FPN, our experiments show that the proposed BorderRPN gains 3.6 $AP$ improvement.

\setlength{\tabcolsep}{3.5pt}
\begin{table}[t]
\caption{Results of combining BorderDet with one-stage detector~(RetinaNet) and two-stage detector~(Faster R-CNN, FPN based)}
\small
\centering
\begin{tabular}{l|cccccc}
\toprule
Method & $AP$ & $AP_{50}$ & $AP_{75}$ & $AP_{S}$ & $AP_{M}$ & $AP_{L}$ \\
\midrule
Retinanet~\cite{focal_loss} & 36.1 & 55.0 & 38.4 & 19.1 & 39.6 & 48.2\\
BD-Retinanet                & \textbf{38.4} & 56.5 & 55.5 & 22.4 & 41.6 & 51.0\\
\midrule
FPN~\cite{faster_rcnn}      & 37.1 & 58.7 & 40.3 & 21.1 & 40.3 & 48.6\\
BD-FPN                      & \textbf{40.7} & 57.8 & 44.3 & 21.9 & 43.7 & 54.8\\
\bottomrule
\end{tabular}
\setlength{\tabcolsep}{12pt}
\label{tab:generalization of BorderDet}
\end{table}

\subsection{Comparisons with State-of-the-art Detectors} The BorderDet, based on FCOS and ResNet-101 backbone, is compared to the state-of-the-art methods in Table~\ref{tab:sota} under standard setting and advanced setting. The standard setting is the same as the setting in Sec.~\ref{sec:exp_setting}. The advanced setting follows the setting that using the jitter over scales \{640, 672, 704, 736, 768, 800\}, and number the training iterations are doubled to 180K. Table.~\ref{tab:sota} shows the comparison with the state-of-the-art detectors on the MS-COCO test-dev set. With the standard setting, the proposed BorderDet achieves an AP of 43.2. It surpasses the anchor-free approaches including GuidedAnchoring, FSAF and CornerNet. By adopting advanced settings, BorderDet reaches 50.3 AP, the state of the art among existing one-stage methods and two-stage methods.

\setlength{\tabcolsep}{1.8pt}
\begin{table}[htb]
\footnotesize
\centering
\caption{BorderDet vs. the state-of-the-art mothods~(single model) on COCO test-dev set. $``\dag''$ indicates multi-scale training. ``$\ddagger$'' indicates the multi-scale testing}
\begin{tabular}{l|c|c|cccccc}
\toprule
Method                              & Backbone                  & Iter.               & AP   & $AP_{50}$ & $AP_{75}$ & $AP_{S}$ & $AP_{M}$ & $AP_{L}$ \\
\midrule
FPN~\cite{fpn}       & ResNet-101-FPN            & 180k          & 36.2 & 59.1 & 39.0 & 18.2     & 39.0     & 48.2      \\
Mask R-CNN~\cite{mask_rcnn}         & ResNet-101-FPN            & 180k          & 38.2 & 60.3 & 41.7 & 20.1     & 41.1     & 50.2     \\
Cascade R-CNN~\cite{cai2018cascade}  & ResNet-101                & 280k          & 42.8 & 62.1 & 46.3 & 23.7     & 45.5      & 55.2          \\
\midrule
RefineDet512~\cite{refinedet}       & Resnet-101                & 280k          & 41.8 & 62.9 & 45.7 & 25.6     & 45.1      & 54.1          \\
RetinaNet~\cite{focal_loss}         & ResNet-101-FPN            & 135k          & 39.1 & 59.1 & 42.3 & 21.8     & 42.7      & 50.2      \\
FSAF~\cite{FSAF}                    & ResNet-101-FPN            & 135k          & 40.9 & 61.5 & 44.0 & 24.0     & 44.2      & 51.3      \\
FCOS~\cite{fcos}                    & ResNet-101-FPN            & 180k          & 41.5 & 60.7 & 45.0 & 24.4     & 44.8      & 51.6      \\
FCOS-imprv~\cite{fcos}              & ResNet-101-FPN            & 180k          & 43.0 & 61.7 & 46.3 & 26.0     & 46.8      & 55.0      \\
CornerNet~\cite{cornernet_law}      & Hourglass-104             & 500k          & 40.6 & 56.4 & 43.2 & 19.1     & 42.8      & 54.3      \\
CenterNet~\cite{duan2019centernet}  & Hourglass-104             & 500k          & 44.9 & 62.4 & 48.1 & 25.6     & 47.4      & 57.4       \\
\midrule
BorderDet & ResNet-101-FPN        & 90k  & 43.2 & 62.1 & 46.7 & 24.4 & 46.3 & 54.9 \\
BorderDet$\dag$ & ResNet-101-FPN    & 180k & 45.4 & 64.1 & 48.8 & 26.7 & 48.3 & 56.5 \\
BorderDet$\dag$ & ResNeXt-64x4d-101 & 180k & 46.5 & 65.7 & 50.5 & 29.1 & 49.4 & 57.5 \\
BorderDet$\dag$ & ResNet-101-DCN    & 180k & 47.2 & 66.1 & 51.0 & 28.1 & 50.2 & 59.9 \\
BorderDet$\dag$ & ResNeXt-64x4d-101-DCN & 180k & 48.0 & 67.1 & 52.1 & 29.4 & 50.7 & 60.5 \\
BorderDet$\ddagger$ & ResNeXt-64x4d-101-DCN & 180k & 50.3 & 68.9 & 55.2 & 32.8 & 52.8 & 62.3 \\
\bottomrule
\end{tabular}
\label{tab:sota}
\end{table}

\section{Conclusion}
In this work, we present the BorderDet, a simple yet effective network architecture that extracts border features in both the classification and regression procedure to improve the localization ability of the object detector. The introduced border features are extracted by a novel operation called BorderAlign. Through BorderAlign, the object detector is able to adaptively learn to extract the features of the extreme point on each borders. Extensive experiments are conducted to validate the BorderAlign has higher performance than the previous feature refinement operations.


\clearpage
%
%
\bibliographystyle{splncs04}
\bibliography{egbib}
\end{document}